\documentclass[runningheads]{llncs}

\usepackage[T1]{fontenc}
\usepackage{graphicx}
\usepackage{amssymb}
\usepackage{hyperref}
\usepackage{dsfont}
\usepackage{epstopdf}

\usepackage{amsmath}
\usepackage{multirow}
\usepackage{soul}
\usepackage{booktabs}
\usepackage{bbding}
\usepackage{makecell}
\usepackage{stfloats}
\usepackage{xcolor}
\usepackage{makecell}
\usepackage{xspace}
\usepackage{url}
\usepackage{wrapfig}
\usepackage{multirow}
\usepackage{makecell}
\usepackage{booktabs, colortbl}
\usepackage{longtable}
\usepackage{hhline}
\usepackage{algpseudocode}
\usepackage{listings}
\usepackage[ruled]{algorithm2e}
\definecolor{myblue}{HTML}{0072C6}
\definecolor{myyellow}{HTML}{FFFADF}
\definecolor{myred}{HTML}{FF0000}
\usepackage[misc]{ifsym} 

\usepackage{floatrow}
\newfloatcommand{capbtabbox}{table}[][\FBwidth]
\newcommand{\para}[1]{\vspace{.05in}\noindent\textbf{#1}}

\begin{document}
\title{SegMamba: Long-range Sequential Modeling Mamba For 3D Medical Image Segmentation}

\titlerunning{SegMamba}
%

\author{Zhaohu Xing\inst{1} \and
Tian Ye\inst{1} \and
Yijun Yang\inst{1} \and Guang Liu\inst{2} 
\and Lei Zhu\inst{1,3} \textsuperscript{(\Letter)}
}

\institute{The Hong Kong University of Science and Technology (Guangzhou) \and
Beijing Academy of Artificial Intelligence \and
The Hong Kong University of Science and Technology \\
\email{leizhu@ust.hk}
}

\maketitle              
\begin{abstract}
The Transformer architecture has demonstrated remarkable results in 3D medical image segmentation due to its capability of modeling global relationships.
However, it poses a significant computational burden when processing high-dimensional medical images.
Mamba, as a State Space Model (SSM), has recently emerged as a notable approach for modeling long-range dependencies in sequential data, and has excelled in the field of natural language processing with its remarkable memory efficiency and computational speed.
Inspired by this, we devise \textbf{SegMamba}, a novel 3D medical image \textbf{Seg}mentation \textbf{Mamba} model, to effectively capture long-range dependencies within whole-volume features at every scale.
Our SegMamba outperforms Transformer-based methods in whole-volume feature modeling, maintaining high efficiency even at a resolution of {$64\times 64\times 64$}, where the sequential length is approximately 260k.
Moreover, we collect and annotate a novel large-scale dataset (named CRC-500) to facilitate benchmarking evaluation in 3D colorectal cancer (CRC) segmentation.
%
Experimental results on our CRC-500 and two public benchmark datasets further demonstrate the effectiveness and universality of our method.
The code for SegMamba is publicly available at:
\href{https://github.com/ge-xing/SegMamba}{https://github.com/ge-xing/SegMamba}.

\end{abstract}

\section{Introduction}

%
3D medical image segmentation plays a vital role in
computer-aided diagnosis.
Accurate segmentation results can alleviate the diagnostic burden on doctors for various diseases.
To improve segmentation performance, extending model's receptive field within 3D space is a critical aspect.
The large-kernel convolution layer~\cite{luo2023lkd} is proposed to model a broader range of features.
3D UX-Net~\cite{lee20223d} introduces a new architecture that utilizes the convolution layer with a large kernel size ($7\times 7\times 7$) as the basic block to facilitate larger receptive fields.
However, CNN-based methods struggle to model global relationships due to the inherent locality of the convolution layer.

Recently, the Transformer architecture~\cite{vaswani2017attention,xing2022nestedformer,yang2024vivim,xing2024hybrid,wang2023video}, utilizing a self-attention module to extract global information, has been extensively explored for 3D medical image segmentation.
%
%
For instance, UNETR~\cite{hatamizadeh2022unetr} employs the Vision Transformer (ViT)~\cite{dosovitskiy2020image} as its encoder to learn global information in a single-scale sequence.
%
SwinUNETR~\cite{hatamizadeh2022swin} leverages the Swin Transformer~\cite{liu2021swin} as the encoder to extract multi-scale features. 
%
While these transformer-based methods improve the segmentation performance, they introduce significant computational costs because of the quadratic complexity in self-attention.

To overcome the challenges of long sequence modeling, Mamba~\cite{gu2023mamba,liu2024vmamba}, which originates from state space models (SSMs)~\cite{kalman1960new}, is designed to model long-range dependencies and enhance the efficiency of training and inference through a selection mechanism and a hardware-aware algorithm.
%
%
U-Mamba~\cite{ma2024u} integrates the Mamba layer into the encoder of nnUNet~\cite{isensee2021nnu} to enhance general medical image segmentation.
Meanwhile, Vision Mamba~\cite{zhu2024vision} introduces the Vim block, which incorporates bidirectional SSM for global visual context modeling.
%
%
However, Mamba has not been fully explored in 3D medical image segmentation.

In this paper, we introduce SegMamba, a novel framwork that combines the U-shape structure with Mamba for modeling the whole volume global features at various scales. 
To our knowledge, this is the first method utilizing Mamba specifically for 3D medical image segmentation.
To enhance the whole-volume sequential modeling of 3D features, we design a tri-orientated Mamba (ToM) module.
Subsequently, we further design a gated spatial convolution (GSC) module to enhance the spatial feature representation before each ToM module.
Furthermore, we design a feature-level uncertainty estimation (FUE) module to filter the multi-scale features from encoder, enabling improved feature reuse.
%
%
Finally, we propose a new large-scale dataset for 3D colorectal cancer segmentation named CRC-500, which consists of 500 3D computed tomography (CT) scans with expert annotations.
%
%
%
Extensive experiments are conducted on three datasets, demonstrating the effectiveness and universality of our method.
SegMamba exhibits a remarkable capability to model long-range dependencies within volumetric data,  while maintaining outstanding inference efficiency.

\vspace{-1mm}
\section{Method}
\vspace{-1mm}

\begin{figure*}[!t]
\centering
\includegraphics[width=0.95\textwidth]{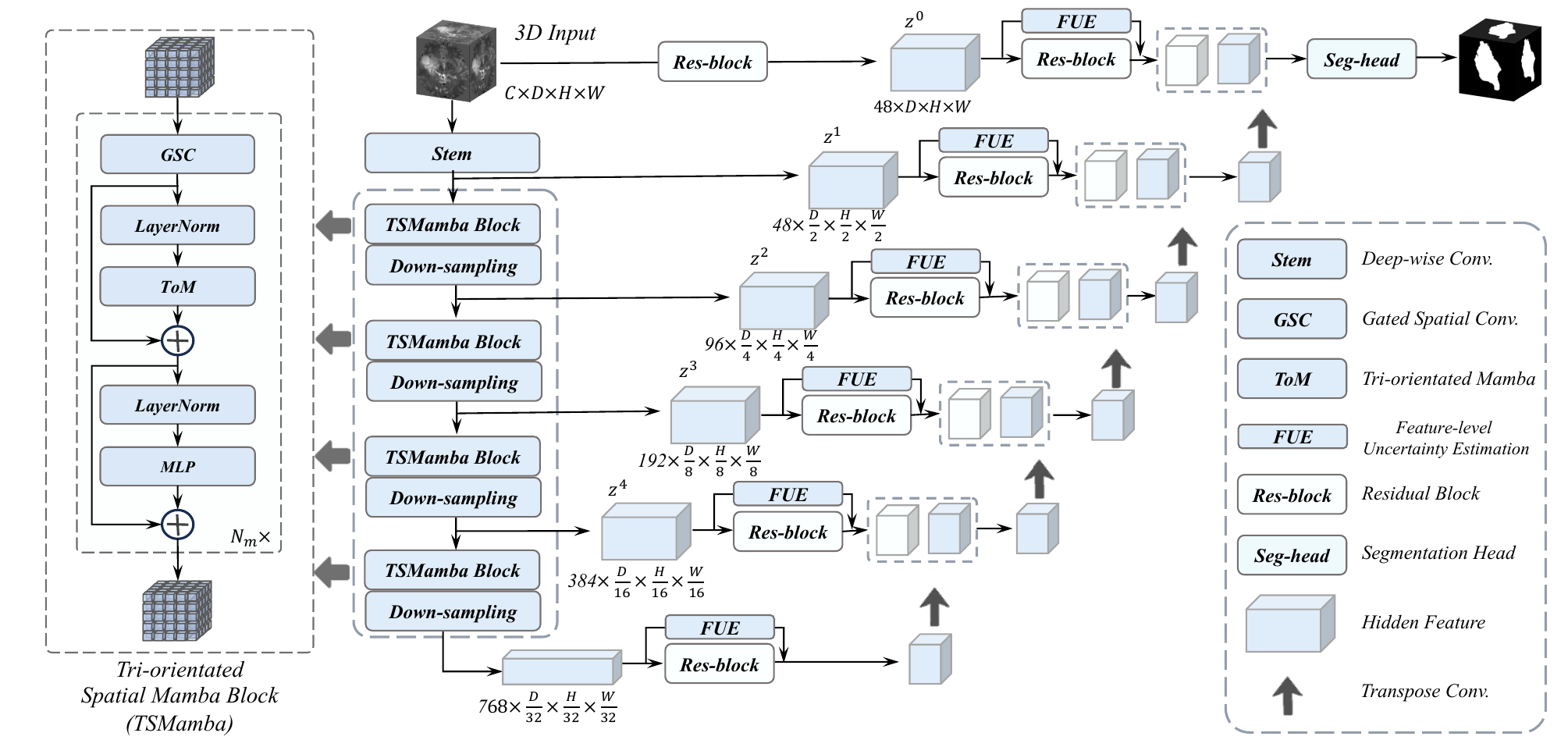}
\label{fig:method}
\caption{
An overview of the proposed SegMamba.
The encoder comprises a stem layer and multiple TSMamba blocks designed to extract multi-scale features. 
Within each TSMamba block, a gated spatial convolution (GSC) module models the spatial features, and a tri-orientated Mamba (ToM) module represents global information from various directions. 
Furthermore, we develop a feature-level uncertainty estimation (FUE) module to filter multi-scale features, facilitating more robust feature reuse.
} 
\end{figure*}

SegMamba mainly consists of three components: 1) a 3D feature encoder with multiple tri-orientated spatial Mamba (TSMamba) blocks for modeling global information at different scales, 2) a 3D decoder based on the convolution layer for predicting segmentation results, and 3) skip-connections with feature-level uncertainty estimation (FUE) for feature enhancement.
Fig. \ref{fig:method} illustrates the overview of the proposed SegMamba.
We further describe the details of the encoder and decoder in this section.

\vspace{-1mm}
\subsection{Tri-orientated Spatial Mamba (TSMamba) Block }

Modeling global features and multi-scale features is critically important for 3D medical image segmentation.
Transformer architectures can extract global information, but it incurs a significant computational burden when dealing with overly long feature sequences.
To reduce the sequence length, methods based on Transformer architectures, such as UNETR, directly down-sample the 3D input with a resolution of $D\times H\times W$ to $\frac{D}{16}\times \frac{H}{16}\times \frac{W}{16}$.
However, this approach limits the ability to encode multi-scale features, which are essential for predicting segmentation results via the decoder.
To overcome this limitation, we design a TSMamba block to enable both multi-scale and global feature modeling while maintains a high efficiency during training and inference.

As illustrated in Fig. \ref{fig:method}, the encoder consists of a stem layer and multiple TSMamba blocks.
For the stem layer, we employ a depth-wise convolution with a large kernel size of $7\times 7\times 7$, with a padding of $3\times 3\times 3$, and a stride of $2\times 2\times 2$.
Given a 3D input volume $I \in \mathbb{R}^{C\times D\times H\times W}$, where $C$ denotes the number of input channels, the first scale feature $z_0 \in \mathbb{R}^{48\times \frac{D}{2}\times \frac{H}{2}\times \frac{W}{2}}$ is extracted by the stem layer.
Then, $z_0$ is fed through each TSMamba block and corresponding down-sampling layers. 
For the $m^{th}$ TSMamba block, the computation process can be defined as:
\begin{equation}
\begin{aligned}
&
\hat{z}^l_{m}=GSC({z}^l_{m}),  \quad
\tilde{z}^l_m={ToM}\left(\operatorname{LN}\left(\hat{z}^{l}_{m}\right)\right)+\hat{z}^{l}_{m}, \quad
z^{l+1}_m=\operatorname{MLP}\left(\operatorname{LN}\left(\tilde{z}^l_{m}\right)\right)+\tilde{z}^l_{m},
\end{aligned}
\end{equation}
where the $GSC$ and $ToM$ denote the proposed gated spatial convolution module and tri-orientated Mamba module, respectively, which will be discussed next. 
$l\in \{0, 1,...,N_m-1\}$, $\operatorname{LN}$ denotes the layer normalization, and $\operatorname{MLP}$ represents the multiple layers perception layer to enrich the feature representation.

\begin{figure*}[!t]
\centering
\includegraphics[width=0.9\textwidth]{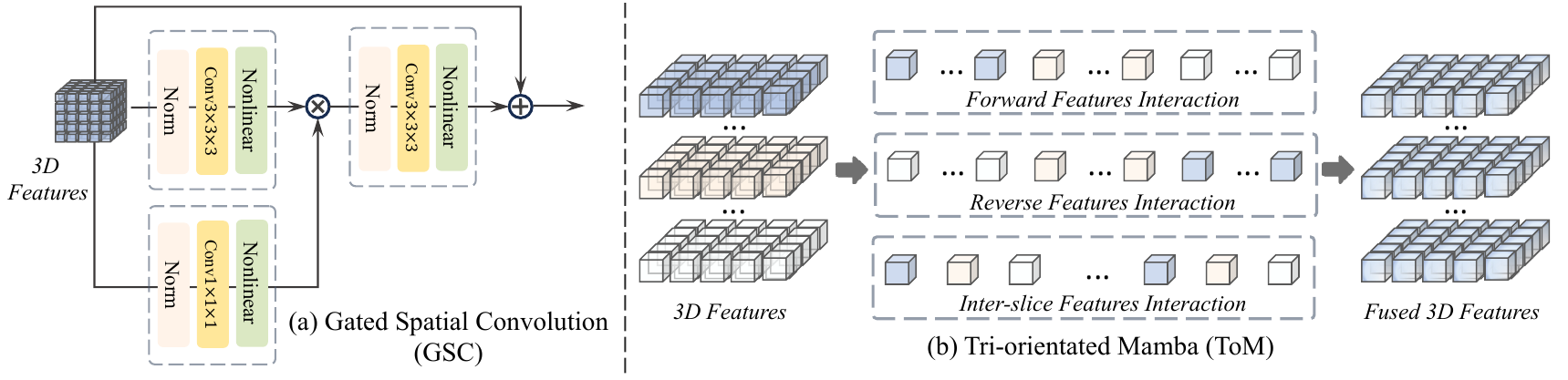}
\caption{
(a) The gated spatial convolution. (b) The tri-orientated Mamba.
} 
\label{fig:modules}

\end{figure*}

\para{Gated Spatial Convolution (GSC)}
The Mamba layer models feature dependencies by flattening 3D features into a 1D sequence, which lacks spatial information. 
Therefore, to capture the spatial relationships before the Mamba layer, we have designed a gated spatial convolution (GSC) module.
As shown in Fig. \ref{fig:modules} (a), the input 3D features are fed into two convolution blocks (each convolution block contains a norm, a convolution, and a nonlinear layer), with the convolution kernel sizes being $3\times 3\times 3$ and $1\times 1\times 1$.
Then these two features are multiplied pixel-by-pixel to control the information transmission similar to the gate mechanism~\cite{liu2021pay}.
Finally, a convolution block is used to further fuse the features, while a residual connection is utilized to reuse the input features.
\begin{equation}
    GSC(z) = z + C^{3\times 3\times 3}(C^{3\times 3\times 3}(z) \cdot C^{1\times 1\times 1}(z)),
\end{equation}
where $z$ denotes the input 3D features and $C$ denotes the convolution block.

\para{Tri-orientated Mamba (ToM)}
The original Mamba block models global dependencies in one direction, which does not suit high-dimensional medical images.
Therefore, in the TSMamba block, to effectively model the global information of high-dimensional features, we design a tri-orientated Mamba module that computes the feature dependencies from three directions. 
As shown in Fig. \ref{fig:modules} (b), we flatten the 3D input features into three sequences to perform the corresponding feature interactions and obtain the fused 3D features.
\begin{equation}
    ToM(z) = Mamba(z_f) + Mamba(z_r) + Mamba(z_s),
\end{equation}
where $Mamba$ represents the Mamba layer used to model the global information within a sequence. The symbol $f$, $r$, $s$ denote flattening in the forward direction, reverse direction, and inter-slice direction, respectively.

\vspace{-1mm}
\subsection{Feature-level Uncertainty Estimation (FUE)}
\vspace{-1mm}
The multi-scale features from the encoder include uncertainty information~\cite{zhao2023uncertainty,xing2023diff} for various structures, such as background and tumor, in 3D data.
To enhance features with lower uncertainty across multiple scales, we design a simple feature-level uncertainty estimation (FUE) module within the skip connections. 
As illustrated in Fig. \ref{fig:method}, for the $i^{th}$ scale feature $z^i \in \mathbb{R}^{C^i\times D^i\times H^i\times W^i}$, we calculate the mean value across the channel dimension and then use a sigmoid function $\sigma$ to normalize this feature.
The computation process of the uncertainty $u^i$ can be summarized as follows:
\begin{equation}
\label{Eq:fun}
    u^i = -\bar{z}^i log(\bar{z}^i) ,~ \text{where} ~\bar{z}^i = \sigma( \frac{1}{C^i}\sum_{c=1}^{C^i}{z_c^i}).
\end{equation}
Hence, the final $i^{th}$ scale feature is represented as $\tilde{z}^i=z^i + z^i \cdot (1 - u^i)$.

\section{Experiments}

\subsection{Collected Colorectal Cancer Segmentation Dataset (CRC-500)}

\begin{figure*}[!t]
\centering
\includegraphics[width=0.8\textwidth]{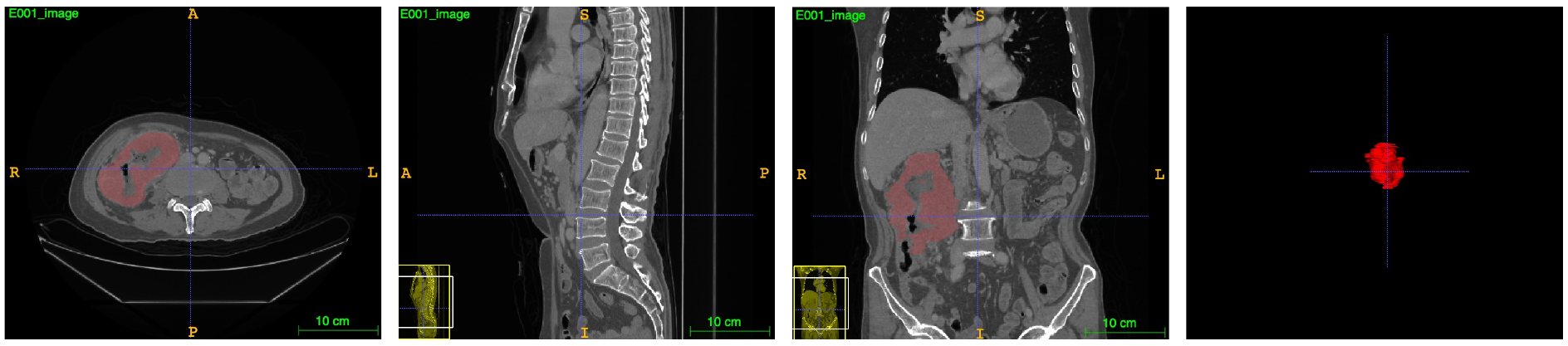}
\caption{
The data visualization for CRC-500 dataset.
} 
\label{fig:data_example}
\end{figure*}

Colorectal cancer (CRC) is the third most common cancer worldwide among men and women, the second leading cause of
death related to cancer, and the primary cause of death in gastrointestinal cancer~\cite{granados2017colorectal}.
%
%
However, as shown in Table \ref{tab:related_datasets}, the existing 3D colorectal cancer segmentation datasets are limited in size, and most of them are private.
%
%
%
We contribute a new large-scale dataset (named CRC-500), which consists of 500 3D colorectal volumes with corresponding precise annotations from ex
perts. 
Fig. \ref{fig:data_example} presents examples in 2D format from our proposed CRC-500 dataset. 

\begin{table*}[!t]
    \centering
    \caption{Comparison between related datasets and our CRC-500 dataset.} 
    \label{tab:related_datasets}
    \renewcommand\arraystretch{1.2}
    \setlength\tabcolsep{5pt}
    \resizebox{0.8\textwidth}{!}{
    \begin{tabular}{c | c | c | c }
    \toprule

    {Related Datasets} & {Rectal Cancer} & {Colon Cancer}  & {Volume Number}  \\
    \midrule
    
    3D RU-Net~\cite{myronenko20183d} & \Checkmark & \Checkmark & 64  \\
    MSDenseNet~\cite{lee20223d} & \Checkmark & \Checkmark & 43   \\
    MSD~\cite{roy2023mednext} & \XSolidBrush & \Checkmark & 190   \\
    Zhang et al.~\cite{hatamizadeh2022unetr} & \Checkmark & \Checkmark & 388  \\

    \midrule
    \rowcolor{gray!15}
    Our CRC-500 & \Checkmark & \Checkmark & 500  \\
    
    \bottomrule
    \end{tabular}
    }
\end{table*}

\para{Dataset Construct}
The CT scans were acquired from January 2008 to April 2020.
All sensitive patient information has been removed.
Each volume was annotated by a professional doctor and calibrated by another professional doctor.
%

\para{Dataset Analysis}
All the CT scans share the same in-plane dimension of $512\times 512$, and the dimension along the z-axis ranges from 94 to 238, with a median of 166.
The in-plane spacing ranges from $0.685\times 0.685$ mm to $0.925\times 0.925$ mm, with a median of $0.826\times 0.826$ mm, and the z-axis spacing is from 3.0 mm to 3.75 mm, with a median of 3.75 mm.

\subsection{Public Benchmarks and Implementation}

\para{BraTS2023 Dataset}
The BraTS2023 dataset~\cite{menze2014multimodal,bakas2017advancing,kazerooni2023brain} contains a total of 1,251 3D brain MRI volumes. Each volume includes four modalities (namely T1, T1Gd, T2, T2-FLAIR) and three segmentation targets (WT: Whole Tumor, ET: Enhancing Tumor, TC: Tumor Core).
%

\para{AIIB2023 Dataset}
The AIIB2023 dataset~\cite{nan2023hunting}, the first open challenge and publicly available dataset for airway segmentation.
The released data include 120 high-resolution computerized tomography scans with precise expert annotations, providing the first airway reference for fibrotic lung disease.



\begin{table*}[!t]
    \centering
    \caption{Quantitative comparison on the BraTS2023 and AIIB2023 datasets. The bold value indicates the best performance.} 

    \label{tab:brats23}
    \renewcommand\arraystretch{1.2}
    \resizebox{1\textwidth}{!}{
    \begin{tabular}{c | c c c c c c c c | c c c}
    \toprule

     \multirow{3}{*}{Methods} & \multicolumn{8}{c|}{BraTS2023} & \multicolumn{3}{c}{AIIB2023}\\
     & \multicolumn{2}{c}{WT} & \multicolumn{2}{c}{TC}  & \multicolumn{2}{c}{ET}  & \multicolumn{2}{c|}{Avg} & \multicolumn{3}{c}{Airway Tree}\\
    
    & Dice $\uparrow$ & HD95 $\downarrow$ & Dice $\uparrow$ & HD95 $\downarrow$ & Dice $\uparrow$ & HD95 $\downarrow$ & Dice $\uparrow$ & HD95 $\downarrow$ & IoU $\uparrow$ & DLR $\uparrow$ & DBR $\uparrow$ \\
    \midrule

    SegresNet~\cite{myronenko20183d} & 92.02 & 4.07 & 89.10  & 4.08  & 83.66 & 3.88 & 88.26 & 4.01 & 87.49	& 65.07 & 	53.91 \\
   
    UX-Net~\cite{lee20223d} & 93.13 & 4.56 & 90.03 &5.68 &	85.91 & 4.19  & 	89.69  & 4.81 & 87.55  & 	65.56 &	54.04 \\
    MedNeXt~\cite{roy2023mednext} & 92.41  &4.98 & 	87.75  & 4.67  &	83.96  & 4.51  & 88.04 &4.72 & 85.81 &  57.43 & 47.34  \\
    \midrule

    UNETR~\cite{hatamizadeh2022unetr} & 92.19 & 6.17 & 	86.39  & 5.29  & 84.48  & 5.03 & 87.68  & 5.49 & 83.22 & 	48.03 & 	38.73  \\
    SwinUNETR~\cite{hatamizadeh2022swin} & 92.71  & 5.22 & 	87.79  & 4.42  & 	84.21  & 4.48 & 	88.23 & 4.70 & 87.11  & 63.31   & 	52.15\\
    SwinUNETR-V2~\cite{he2023swinunetr} & 93.35 & 5.01 & 	89.65  & 4.41 & 	85.17 & 4.41  & 	89.39  & 4.51 & 87.51 & 	64.68   & 	53.19 \\

    \midrule
    \rowcolor{gray!15} 
    Our method & \textbf{93.61} & \textbf{3.37 } & \textbf{92.65 } & \textbf{3.85 } & \textbf{87.71} & \textbf{3.48 } & \textbf{91.32 } & \textbf{3.56} & \textbf{88.59} & \textbf{70.21} & \textbf{61.33} \\
    
    \bottomrule
    \end{tabular}
    }
\end{table*}

\begin{table*}[tp]
\small 

\begin{floatrow}
\resizebox{0.45\textwidth}{!}{
\centering
\renewcommand\arraystretch{1.1}
\ttabbox{\caption{Quantitative comparison on the CRC-500 dataset.}}{%
\label{tab:ccr-500}
\begin{tabular}{c c c} 
    \toprule

    Methods & Dice $\uparrow$ & HD95 $\downarrow$ \\
    \midrule
    SegresNet~\cite{myronenko20183d} & 46.10 & 34.97 \\
    UX-Net~\cite{lee20223d} & 45.73 & 49.73 \\
    MedNeXt~\cite{roy2023mednext} & 35.93 & 52.54 \\
    UNETR~\cite{hatamizadeh2022unetr} & 33.70 & 61.51 \\
    SwinUNETR~\cite{hatamizadeh2022swin} & 38.36 & 55.05 \\
    SwinUNETR-V2~\cite{he2023swinunetr} & 41.76 & 58.05 \\
    \midrule
    \rowcolor{gray!15} 
    Our method & \textbf{48.46} & \textbf{28.52} \\
    \bottomrule
    \end{tabular}
    }
}

\resizebox{0.45\textwidth}{!}{
\centering
\renewcommand\arraystretch{1.4}
\begin{floatrow}
\ttabbox{\caption{Ablation study for different modules on the CRC-500 dataset.}
\label{tab:ablation_result}}{%
\begin{tabular}{c c c c c c} 
    \toprule
    
    Methods & \multicolumn{3}{c}{Modules} & \multirow{2}{*}{Dice $\uparrow$} & \multirow{2}{*}{HD95 $\downarrow$} \\
    \cmidrule{2-4}
     & GSC & ToM & FUE &  &  \\
    \midrule
    
    M1 &   &  & & 45.34 & 43.01 \\
    M2 & \Checkmark &  &  & 46.65 & 37.01 \\
    M3  &  & \Checkmark &  & 47.22 & 33.32 \\
    M4  & \Checkmark & \Checkmark &  & 48.02 & 30.89 \\
    \midrule
    \rowcolor{gray!15} 
    Our method  & \Checkmark & \Checkmark & \Checkmark & \textbf{48.46} & \textbf{28.52} \\
   
    \bottomrule
    \end{tabular}
    }
\end{floatrow}
}
\end{floatrow}
\end{table*}

\para{Implementation Details}
Our model is implemented in PyTorch 2.0.1-cuda11.7 and Monai 1.2.0.
During training, we use a random crop size of $128\times128\times128$ and a batch size of 2 per GPU for each dataset. 
%
%
We employ cross-entropy loss across all experiments and utilize an SGD optimizer with a polynomial learning rate scheduler (initial learning rate of 1e-2, a decay of 1e-5).
We run 1000 epochs for all datasets and adopt the following data augmentations: additive brightness, gamma, rotation, scaling, mirror, and elastic deformation.
All experiments are conducted on a cloud computing platform with four NVIDIA A100 GPUs.
%
For each dataset, we randomly allocate 70\% of the 3D volumes for training, 10\% for validation, and the remaining 20\% for testing.

\subsection{Comparison with SOTA Methods}

\begin{figure*}[!t]
\centering
\includegraphics[width=0.95\textwidth]{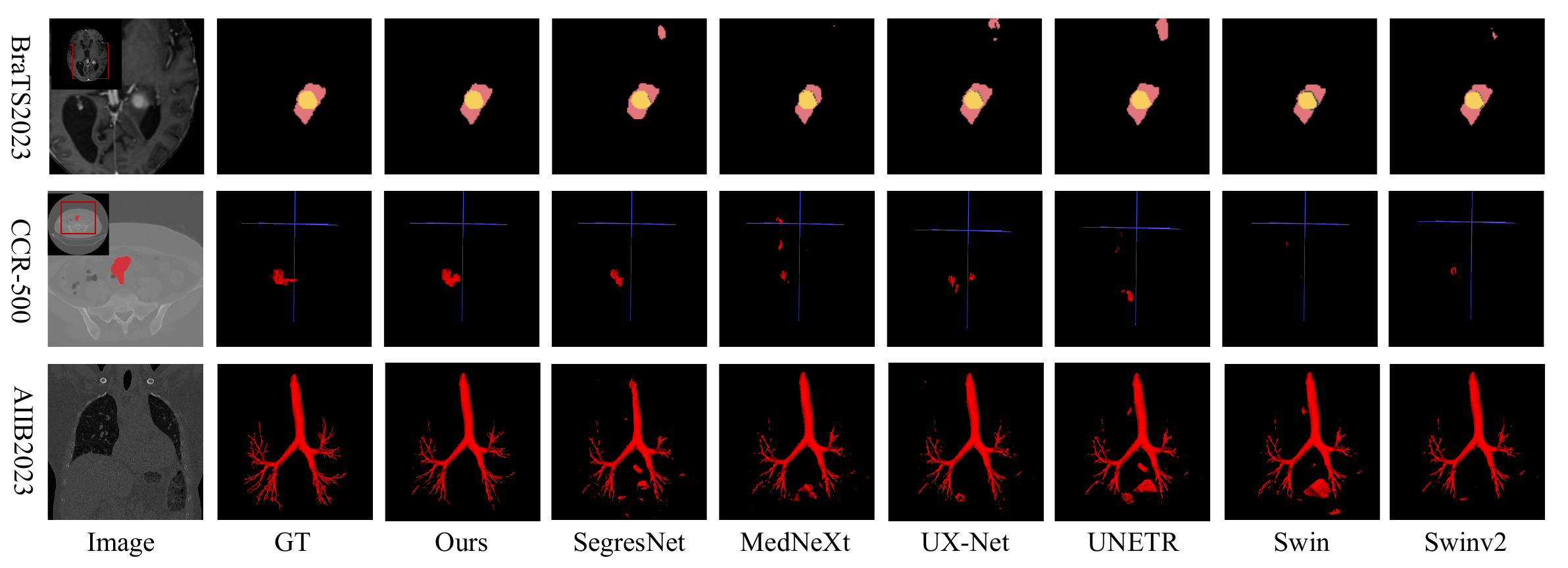}
\caption{
Visual comparisons of proposed SegMamba and other state-of-the-art methods. Swin denotes SwinUNETR and Swinv2 denotes SwinUNETR-V2.
} 
\label{fig:visual_comparison}

\vspace{-1mm}
\end{figure*}

\begin{table*}[!t]
    \centering
    \caption{Ablation study for different global modeling modules. TM denotes training memory, IM denotes inference memory, IT denotes inference time, and OOM represents out of memory.} 
    \label{tab:merits}
    \renewcommand\arraystretch{1.2}
    \setlength\tabcolsep{2pt}
    \resizebox{0.90\textwidth}{!}{
    \begin{tabular}{c c c c c c c c}
    \toprule
    Methods & \makecell{Core\\ module} & \makecell{Input \\ resolution} & \makecell{Sequence\\ length} & \makecell{TM\\ (M)} & \makecell{IM \\ (M)} & \makecell{IT\\ (case/s)} & Is Global \\
    \midrule

    M5 & \makecell{Large-kernel\\ convolution} & $128^3$ & 262144 & 18852 & 5776 & 1.92  & \XSolidBrush  \\

     M6 & \makecell{SwinTransformer}  & $128^3$ & 262144  & 34000  & 9480  & 1.68 & \XSolidBrush \\

    M7  & Self-attention  & $128^3$ & 262144  & OOM  & -  & - & 	\Checkmark\\
    \midrule
    
    \rowcolor{gray!15} 
    Our method & TSMamba & $128^3$ & 262144 & \textbf{17976} & 6279 & \textbf{1.51} & \Checkmark \\
    
    \bottomrule
    \end{tabular}
    }
\end{table*}

We compare SegMamba with six state-of-the-art segmentation methods, including three CNN-based methods (SegresNet~\cite{myronenko20183d},  UX-Net{~\cite{lee20223d}}, MedNeXt~\cite{roy2023mednext}), and three transformer-based methods ( UNETR~\cite{hatamizadeh2022unetr}, SwinUNETR~\cite{hatamizadeh2022swin}, and SwinUNETR-V2~\cite{he2023swinunetr}). 
%
%
For a fair comparison, we utilize public implementations of
these methods to retrain their networks under the same settings.
%
The Dice score (Dice) and 95\% Hausdorff Distance (HD95) are adopted for quantitative comparison on the BraTS2023 and CCR-500 datasets.
Following \cite{nan2023hunting}, the Intersection over union (IoU), Detected length ratio (DLR), and Detected branch ratio (DBR) are adopted on the AIIB2023 dataset. 

\para{BraTS2023} \
The segmentation results of gliomas for the BraTS2023 dataset are listed in Table \ref{tab:brats23}.
UX-Net, a CNN-based method, achieves the best performance among the comparison methods, with an average Dice of 89.69\% and an average HD95 of 4.81.
%
%
In comparison, our SegMamba achieves the highest Dices of 93.61\%, 92.65\%, and 87.71\%, and HD95s of 3.37, 3.85, and 3.48 on WT, TC, and ET, respectively, showing better segmentation robustness.
%

\para{AIIB2023} 
For this dataset, the segmentation target is the airway tree, which includes many tiny branches and poses challenges in obtaining robust results. As shown in Table \ref{tab:brats23}, our SegMamba achieves the highest IoU, DLR, and DBR scores of 88.59\%, 70.21\%, and 61.33\%, respectively. This also indicates that our SegMamba exhibits better segmentation continuity compared to other methods.
%

\para{CRC-500}
The results on the CRC-500 dataset are listed in Table \ref{tab:ccr-500}. In this dataset, the cancer region is typically small; however, our SegMamba can accurately detect the cancer region and report the best Dice and HD95 scores of 48.46\% and 28.52, respectively.

\para{Visual Comparisons}
To compare the segmentation results of different methods more intuitively, we choose six comparative methods for visual comparison on three datasets.
As depicted in Fig. \ref{fig:visual_comparison}, our SegMamba can accurately detect the boundary of each tumor region on BraTS2023 dataset.
Similar to BraTS2023 dataset, our method accurately detects the cancer region on CRC-500 dataset. The segmentation results show better consistency compared to other state-of-the-art methods.
Finally, on AIIB2023 dataset, our SegMamba can detect a greater number of branches in the airway and achieve better continuity. 

\vspace{-1mm}
\subsection{Ablation Study}

\para{The Effectiveness of Proposed Modules}
As shown in Table \ref{tab:ablation_result}, M1 represents our basic method, which includes only the original Mamba layer.
In M2, we introduce our GSC module. Compared to M1, M2 achieves an improvement of 2.88\% and 13.95\% in Dice and HD95.
This shows that the GSC module can improve the spatial representation before the ToM module.
Then, in M3, we introduce the ToM module, which model the global information from three directions.
M3 reports the Dice and HD95 of 47.22\% and 33.32, with an improvement of 1.22\% and 9.97\% compared to M2. 
Furthermore, we introduce the GSC and ToM modules simultaneously, resulting in an increase of 1.69\% in Dice and 7.29\% in HD95.
Finally, our SegMamba introduce both GSC, ToM, and FUE modules, achieving the state-of-the-art performance, with the Dice and HD95 of 48.46\% and 28.52.

\para{The High Efficiency of SegMamba}
We verify the high efficiency of our SegMamba through an ablation study presented in Table \ref{tab:merits}.
M4 is UX-Net~\cite{lee20223d}, which utilizes large-kernel convolution as its core module. 
M5 is SwinUNETR~\cite{hatamizadeh2022swin}, which uses the SwinTransformer as its core module.
Both improve receptive field by computing long range pixels, but they cannot compute the relationship within a global range.
In M6, we use self-attention, a global modeling layer, as the core module, but it is infeasible due to the computational burden.
In comparison, our method uses a Mamba-based global modeling module (TSMamba), and achieves a better training memory (TM) and inference time (IT), even though the maximum flattened sequence length reaches 260k.

\vspace{-1mm}
\section{Conclusion}
In this paper, we propose the first general 3D medical image segmentation method based on the Mamba, called SegMamba.
First, we design a tri-orientated Mamba (ToM) module to enhance the sequential modeling for 3D features. 
To effectively model the spatial relationships before the ToM module, we further design a gated spatial convolution (GSC) module. 
%
Moreover, we design a feature-level uncertainty estimation (FUE) module to enhance the multi-scale features in skip-connections.
Finally, we present a new large-scale dataset
for 3D colorectal cancer segmentation, named CRC-500, to support related research. 
SegMamba exhibits a remarkable capability in modeling long-range dependencies within volumetric data, while maintaining outstanding inference efficiency.
%
Extensive experiments demonstrate the effectiveness and universality of our method.

\subsubsection{Acknowledgments} 
This work is supported by the Guangzhou-HKUST(GZ) Joint Funding Program (No. 2023A03J0671), the Guangzhou Municipal Science and Technology Project (Grant No. 2023A03J0671), and the InnoHK funding launched by Innovation and Technology Commission, Hong Kong SAR.

\subsubsection{Disclosure of Interests}
The authors declare that they have no competing interests.

\bibliographystyle{splncs04}
\bibliography{Paper-0663}

\end{document}